\documentclass{SCIS2019}
\usepackage{makecell}
\usepackage{caption}
\usepackage{float}
\usepackage{graphicx}

\begin{document}
\ArticleType{LETTER}
\Year{2022}
\Month{}
\Vol{}
\No{}
\DOI{}
\ArtNo{}
\ReceiveDate{}
\ReviseDate{}
\AcceptDate{}
\OnlineDate{}

\title{LAMDA-SSL: A Comprehensive Semi-Supervised Learning Toolkit}{LAMDA-SSL: A Comprehensive Semi-Supervised Learning Toolkit}

\author[]{Lin-Han Jia}{}
\author[]{Lan-Zhe Guo}{{guolz@lamda.nju.edu.cn}}
\author[]{Zhi Zhou}{}
\author[]{Yu-Feng Li}{{liyf@lamda.nju.edu.cn}}

\AuthorMark{Lin-Han Jia}

\AuthorCitation{Lin-Han Jia, Lan-Zhe Guo, Zhi Zhou and Yu-Feng Li}

\address[]{National Key Laboratory for Novel Software Technology, Nanjing University, Nanjing 210023, China}

\maketitle

\begin{multicols}{2}
\deareditor

Machine learning, particularly deep learning, has achieved remarkable success across a wide range of tasks. However, most of these tasks demand a substantial amount of labeled training data, which can be challenging to obtain in many real-world applications due to difficulties, costs, or time constraints associated with labeling. In this context, semi-supervised learning (SSL) has emerged as a promising paradigm to alleviate the dependency on large labeled datasets by harnessing the power of unlabeled data~\cite{guo2020safe, li2021towards, guo2022robust}. 

With the increasing interest of the machine learning community in SSL, it is crucial to provide a comprehensive and user-friendly toolkit. However, to the best of our knowledge, there are only two widely used SSL toolkits: the SSL module of ``scikit-learn''~\cite{pedregosa2011scikit}, that focuses on statistical SSL, and ``USB''~\cite{wang2022usb}, a ``Pytorch''~\cite{paszke2019pytorch}-based toolkit that focuses on deep SSL. Unfortunately, these SSL toolkits are not comprehensive enough, as ``scikit-learn'' only includes 3 statistical SSL algorithms, while ``USB'' includes 14 deep SSL algorithms but is limited to classification tasks.

\begin{figure*}[t]
	\centering
    \includegraphics[scale=0.3]{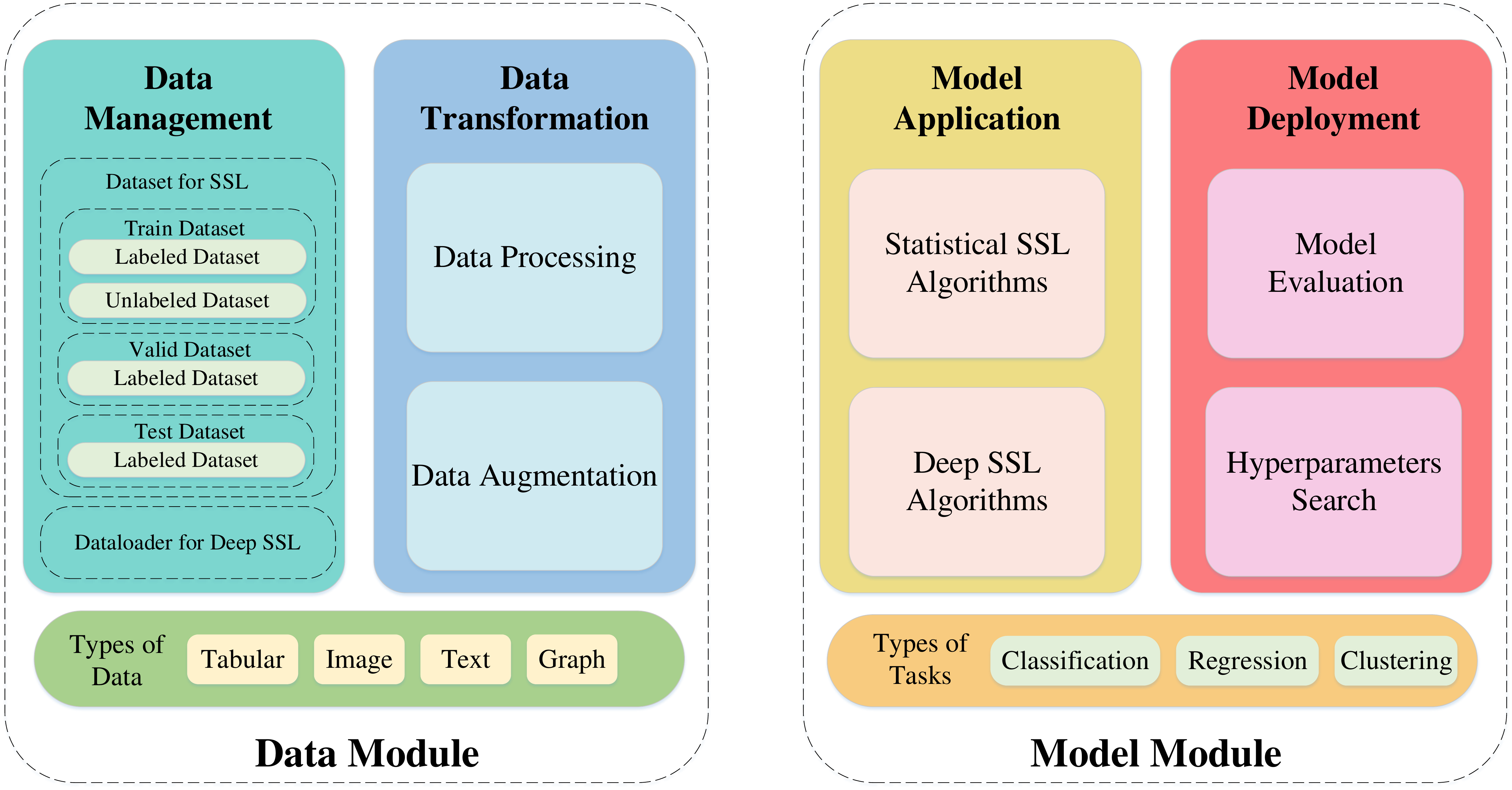}
	\caption{An overview of the design, functions and contents of LAMDA-SSL}
	\label{fig:overview}
\end{figure*}

In this paper, we introduce LAMDA-SSL, an open-sourced Python toolkit for SSL. LAMDA-SSL incorporates both statistical and deep SSL algorithms within a single framework, and it is compatible with popular toolkits such as ``scikit-learn'' and ``Pytorch". Currently, LAMDA-SSL offers a total of 30 SSL algorithms, including 12 statistical and 18 deep SSL algorithms. The toolkit is designed with two main components: data and model (as shown in Figure~\ref{fig:overview}). In the data module, LAMDA-SSL facilitates data management and transformation for 4 types of data, such as tabular, image, text, and graph. In the model module, LAMDA-SSL supports model application and deployment for 3 tasks such as classification, regression, and clustering. LAMDA-SSL is designed to be user-friendly, providing simple interfaces and well-tuned default parameters for entry-level users, while also offering flexibility for component replacement and customization for professional users.

We conducted a comparison between LAMDA-SSL, the SSL module of ``scikit-learn," and ``USB" (Table~\ref{tbl:comp}). To the best of our knowledge, LAMDA-SSL is the first SSL toolkit that seamlessly integrates statistical SSL and deep SSL algorithms within a unified framework. LAMDA-SSL demonstrates superiority in both statistical SSL and deep SSL domains. In the realm of statistical SSL, LAMDA-SSL is more suitable for SSL applications compared to ``scikit-learn." In the realm of deep SSL, LAMDA-SSL outperforms ``USB" in various aspects, including the number of algorithms, data types, task types, and functionality.

The key advantages of LAMDA-SSL include comprehensive functionality, simple interfaces, and extensive documentation.

\textbf{Comprehensive Functionality.} At present, LAMDA-SSL has implemented 30 SSL algorithms, including 12 statistical SSL algorithms and 18 deep SSL algorithms.

For statistical SSL~\cite{chapelle2006semi}, the algorithms implemented for classification include:
\begin{itemize}
	\item Generative methods: SSGMM.
	\item Semi-supervised SVMs: TSVM, LapSVM.
	\item Graph-based methods: Label Propagation, Label Spreading.
	\item Disagreement-based methods: Co-Training, Tri-Training.
	\item Ensemble methods: SemiBoost, Assemble.
\end{itemize}
The algorithms implemented for regression include CoReg and for clustering include Constrained $K$-Means and Constrained Seed $K$-Means.
    
For deep SSL~\cite{oliver2018realistic}, the algorithms implemented for classification include:
\begin{itemize}
	\item Consistency methods: Ladder Network, $\Pi$-Model, Temporal Ensembling, Mean Teacher, VAT, UDA.
	\item Pseudo-label methods: PL, S4L.
	\item Hybrid methods: ICT, MixMatch, ReMixMatch, FixMatch, FlexMatch.
	\item Deep generative methods: Improved GAN, SSVAE.
	\item Deep graph methods: SDNE, GCN, GAT.
\end{itemize}
We also implemented $\Pi$-Model Reg, Mean Teacher Reg and ICT Reg for the regression task.

\begin{table*}[t]
  \centering
  \caption{The comparison of LAMDA-SSL with related SSL toolkits.}
  \label{tbl:comp}
  \begin{tabular}{c c c c}
  \hline\hline
      Toolkit &  \makecell[c]{scikit-learn}   & USB & LAMDA-SSL\\
      \hline
      $\#$ deep SSL algorithms &0&14&18\\
      \hline
      $\#$ statistical SSL algorithms  &3 &0&12\\
      \hline
      Types of data & \makecell[c]{Tabular\\ Image \\ Text} & \makecell[c]{Image\\Text\\Audio} & \makecell[c]{Tabular\\ Image\\ Text\\ Graph}\\
      \hline
      Types of task &\makecell[c]{Classification\\ Regression\\Clustering} & Classification&\makecell[c]{Classification\\ Regression\\Clustering} \\
      \hline
        Hyper-parameters search &$\checkmark$ & $\times$ & $\checkmark$\\
       \hline
    GPU acceleration 
    &$\times$ & $\checkmark$ &$\checkmark$ \\
    \hline
      Distributed learning & $\times$ & $\checkmark$ & $\checkmark$\\
     \hline
     Documentation 
    & $\checkmark$ & $\checkmark$ & $\checkmark$\\ 
    \hline\hline
  \end{tabular}
  \end{table*}

In addition to its diverse range of algorithms, LAMDA-SSL offers 45 data transformation methods and 16 metrics for model evaluation. It also provides flexible interfaces for component replacement and customization, tailored for professional users. Specifically, for deep SSL, LAMDA-SSL allows users to easily replace and customize modules such as \textbf{Dataset}, \textbf{Dataloader}, \textbf{Sampler}, \textbf{Augmentation}, \textbf{Network}, \textbf{Optimizer} and \textbf{Scheduler}, without affecting other modules. This empowers users to achieve low-code implementation for customized deep SSL algorithms by leveraging LAMDA-SSL's \textbf{DeepModelMixin} component, which provides default processing functions for deep SSL. Furthermore, LAMDA-SSL is designed to be compatible with popular toolkits such as ``scikit-learn" and ``Pytorch", inheriting their mechanisms and functions. Similar to ``scikit-learn", LAMDA-SSL supports the pipeline mechanism and includes hyper-parameter search functionality. And like ``Pytorch", LAMDA-SSL takes advantage of GPU acceleration for training and supports distributed learning.

\textbf{Simple Interface.} The APIs of LAMDA-SSL refer to ``scikit-learn'' and all the learners have two basic methods: \textbf{fit()} and \textbf{predict()}. The only difference from the APIs of ``scikit-learn'' is that the \textbf{fit()} method of LAMDA-SSL needs three data items of $X$, $y$ and $unlabeled\_X$ to be input. For deep SSL algorithms, LAMDA-SSL uses \textbf{DeepModelMixin} component to make the APIs of deep SSL algorithms and statistical SSL algorithms unified. Lots of elaborate examples can be found in the online documentation and the Example module of the source code.

\textbf{Extensive Documentation} LAMDA-SSL is open-sourced on GitHub and its detailed documentation is already available at \url{https://ygzwqzd.github.io/LAMDA-SSL/}. This documentation provides a detailed overview of LAMDA-SSL from various perspectives, organized into four parts. The first part covers the mechanism, features, and functions of LAMDA-SSL. The second part offers abundant examples to illustrate the usage of LAMDA-SSL in detail. The third part introduces all the implemented SSL algorithms in LAMDA-SSL, facilitating users in understanding and selecting the right algorithm. The fourth part showcases the APIs of LAMDA-SSL. This thorough documentation significantly reduces the learning difficulty for users to familiarize themselves with the LAMDA-SSL toolkit and SSL algorithms.

We also evaluate LAMDA-SSL according to several quality standards of open source software.
\begin{itemize}
  \item \textbf{Availability.} LAMDA-SSL depends only on standard open-source toolkits and is usable under many operating systems such as Linux, MacOSX, and Windows. The toolkit can be acquired via Pypi easily using ``pip install LAMDA-SSL". The source code of LAMDA-SSL is available at \url{https://github.com/YGZWQZD/LAMDA-SSL}.
  \item \textbf{Reliability.} We conduct comprehensive evaluations of various algorithms in LAMDA-SSL using multiple benchmark datasets, including CIFAR-10, Cora, Boston, among others. The experimental results are accessible on the homepage. Most of the reproduced algorithms exhibit comparable or improved performance compared to the results reported in the original papers.
  \item \textbf{Openness}. LAMDA-SSL is distributed under the MIT license. Contributions from the community are strongly welcome and easy enough because the documentation provides numerous examples showing how to customize modules of LAMDA-SSL.
\end{itemize}

\textbf{Conclusion.} In this paper, we present an easy-to-use, powerful and open-source toolkit in Python for SSL with comprehensive functionality, simple interfaces, complete documentation, and the best support for algorithms, data types, and tasks compared with related toolkits. Our aim is to facilitate SSL research and applications by addressing the challenge of limited labeled data. Going forward, we plan to incorporate advanced algorithms into LAMDA-SSL and expand its application scope in open environments~\cite{zhou2022open}.

\Acknowledgements{This research was supported by the National Key R\&D Program of China (2022ZD0114803), the National Science Foundation of China (62176118).
}

\end{multicols}
\end{document}